\documentclass[11pt]{article}

\usepackage[final]{acl}

\usepackage{times}
\usepackage{latexsym}
\usepackage[T1]{fontenc}
\usepackage[utf8]{inputenc}
\usepackage{microtype}
\usepackage{inconsolata}
\usepackage{hyperref}
\usepackage{graphicx}
\usepackage{booktabs}
\usepackage{tabularx}
\usepackage{amsmath}
\usepackage{stfloats}
\usepackage{pifont}

\usepackage{tikz}
\usetikzlibrary{shapes.geometric, arrows.meta, positioning, calc, backgrounds, fit}
\usepackage{xcolor}

\definecolor{boxbluebg}{RGB}{242, 246, 253}
\definecolor{boxblueborder}{RGB}{175, 195, 230}

\definecolor{boxredbg}{RGB}{253, 244, 244}
\definecolor{boxredborder}{RGB}{230, 185, 185}

\definecolor{boxyellowbg}{RGB}{254, 249, 236}
\definecolor{boxyellowborder}{RGB}{238, 205, 140}

\definecolor{boxpurplebg}{RGB}{247, 244, 252}
\definecolor{boxpurpleborder}{RGB}{198, 185, 226}

\definecolor{boxgreenbg}{RGB}{241, 248, 243}
\definecolor{boxgreenborder}{RGB}{168, 212, 180}

\definecolor{taggray}{RGB}{110, 110, 110}
\definecolor{arrowgray}{RGB}{90, 95, 105}

\newcommand{\yesmark}{\textcolor{green!55!black}{\ding{51}}}
\newcommand{\nomark}{\textcolor{red!70!black}{\ding{55}}}
\newcommand{\partialmark}{\textcolor{orange!80!black}{\textbf{$\sim$}}}

\title{PhoPunct: A Large-Scale Benchmark of Transformer Architectures for Vietnamese Punctuation Restoration}

\author{
  \textbf{Khoa D. Nguyen\textsuperscript{1,2,4}},
  \textbf{Thai Bao Huynh\textsuperscript{1,3,4}},
  \textbf{Binh T. Nguyen\textsuperscript{1,2,4,*}}
  \\
  \textsuperscript{1}AISIA Research Lab, Ho Chi Minh City, Vietnam
  \\
  \textsuperscript{2}University of Science, Ho Chi Minh City, Vietnam
  \\
  \textsuperscript{3}University of Information Technology, Ho Chi Minh City, Vietnam
  \\
  \textsuperscript{4}Vietnam National University, Ho Chi Minh City, Vietnam
  \\
  \small{
    \texttt{ndkhoa.vietnam@gmail.com},
    \texttt{23520105@gm.uit.edu.vn},
    \texttt{ngtbinh@hcmus.edu.vn}
  }
  \\
  \small{\textsuperscript{*}Corresponding author: \texttt{ngtbinh@hcmus.edu.vn}}
}

\begin{document}
\maketitle
\raggedbottom
\begin{abstract}
Vietnamese punctuation restoration is often evaluated on relatively small benchmarks, limiting conclusions about how architectural choices scale with data volume. We present a controlled benchmark on News and Novels corpora containing 22.1 million labeled tokens (19.9M News and 2.2M Novels). We evaluate 17 architectures formed from four pretrained Transformer backbones and optional bidirectional long short-term memory (Bi-LSTM) and conditional random field (CRF) layers, together with a PhoBERT-large--Bi-LSTM--CRF model, PhoPunct. Every model is trained for eight epochs with the same learning rate and without architecture-specific tuning. Under this fixed budget, adding Bi-LSTM and CRF layers does not reliably improve macro F1: in seven of eight matched domain--backbone comparisons, a plain Transformer classifier matches or outperforms its Bi-LSTM--CRF counterpart. PhoPunct obtains the best News result (61.35), whereas XLM-R with a CRF obtains the best Novels result (50.18), exceeding PhoPunct (48.97). These results show that the value of recurrent and structured-prediction heads depends on their optimization regime and motivate explicit separation between architectural improvements and improvements due to architecture-specific tuning. Our source code is available at: \href{https://github.com/aisia-research-lab/PhoPunct}{https://github.com/aisia-research-lab/PhoPunct}
\end{abstract}

\begin{table*}[t]
  \centering
  \small
  \setlength{\tabcolsep}{4pt}
  \begin{tabularx}{\textwidth}{@{}l l >{\raggedright\arraybackslash}X c c@{}}
    \toprule
    Study & Data scale & Focus & Shared budget & Matched heads \\
    \midrule
    \citet{pham2019crf} & N/A & CRF baseline & -- & \nomark \\
    \citet{pham2020vietnamese} & 22.1M tokens & Seven-label DNN & \nomark & \nomark \\
    \citet{tran2021efficient} & $\sim$742K sentences & Transformer heads & \nomark & \partialmark \\
    \citet{uyen2022capitalization} & 425,350 sentences & Case + punctuation & \nomark & \nomark \\
    \textbf{Ours} & \textbf{22.1M tokens} & \textbf{$4\times4+1$ grid} & \yesmark & \yesmark \\
    \bottomrule
  \end{tabularx}
  \caption{Vietnamese punctuation-restoration studies. \yesmark: criterion satisfied; \nomark: not satisfied or not reported; \partialmark: partial comparison; --: not applicable (columns indicate a shared optimization budget across compared architectures and a matched Plain/Bi-LSTM/CRF/Bi-LSTM--CRF head set, respectively). Data scale is reported in each study's own unit (sentences or tokens) and is not directly comparable across rows. Pham et al. (2020) and Ours report identical token counts because both use the same underlying files.}
  \label{tab:related-comparison}
\end{table*}

\section{Introduction}

Punctuation restoration recovers punctuation marks from unpunctuated text. It is an important post-processing step for automatic speech recognition (ASR), where output is commonly produced without reliable punctuation, and it also supports readability, text normalization, and downstream language processing. The task is typically formulated as word-level sequence labeling: given a sequence of lexical tokens, a model predicts the punctuation mark that follows each token or predicts that no mark is present.

Recent systems use contextual representations from pretrained Transformers and may add recurrent or structured-prediction layers. For Vietnamese, \citet{tran2021efficient} evaluated multilingual BERT, XLM-R, and Vietnamese-specific pretrained encoders on News and Novels data. Related Vietnamese work has also examined an earlier CRF-based approach \citep{pham2019crf}, deep neural punctuation prediction \citep{pham2020vietnamese}, and joint capitalization and punctuation recovery \citep{uyen2022capitalization}. These studies establish the effectiveness of pretrained encoders, but the effect of adding a Bi-LSTM or CRF on top of such encoders remains unclear, especially when architectures are compared under the same optimization budget.

We revisit this question at large token scale, training on 22,137,597 labeled tokens from the public News and Novels corpus released by \citet{pham2020vietnamese} and subsequently used by \citet{tran2021efficient}, excluding synthetic augmentation. We evaluate 17 architectures under an identical eight-epoch schedule and a single learning rate for all parameters. The resulting comparison concerns performance under a fixed, shared optimization budget; it is not intended to estimate the separately tuned optimum of every architecture.

This work makes three connected contributions. First, it establishes a large-scale benchmark on the original News and Novels core splits, comprising 22.1 million labeled tokens. Second, it evaluates four pretrained backbones with Plain, Bi-LSTM, CRF, and Bi-LSTM--CRF prediction heads, together with PhoPunct, a PhoBERT-large--Bi-LSTM--CRF model. Third, it identifies a practically important negative result: under the shared budget, recurrent and structured-prediction layers do not consistently improve macro F1. This finding makes the optimization of newly initialized components a central experimental variable rather than an implementation detail.

\section{Related Work}

Punctuation restoration has progressed from feature-based and recurrent sequence models to fine-tuning pretrained Transformers. Efficient Transformer formulations predict punctuation at selected token positions \citep{courtland2020efficient}, while recurrent models with layer-wise attention illustrate the earlier emphasis on explicitly modeling sequential context \citep{kim2019attention}. Modern systems often retain that intuition by placing a Bi-LSTM or CRF above a contextual encoder. The unresolved question is whether those additional layers remain useful once the encoder already provides deep bidirectional context, particularly when all alternatives receive the same optimization budget.

Vietnamese research reflects this progression. \citet{pham2019crf} established an early CRF formulation on 500 news articles and 100 movie-subtitle transcripts, using a smaller, domain-specific label set---period and comma for News, plus question mark and exclamation mark for movie subtitles---rather than the six-mark inventory used here. \citet{pham2020vietnamese} expanded the task to seven categories with a deep neural network and released the News and Novels corpus this benchmark also trains on---the same files, not merely a related dataset. However, that work did not compare its sequence heads with a Plain-classifier control under the same budget, leaving the head-isolation question unresolved. \citet{tran2021efficient} subsequently used the News and Novels setting and found sub-point gains from adding sequence layers to its strongest backbone; we test whether that pattern persists across four backbone families. \citet{uyen2022capitalization} instead changed the output task through joint capitalization and punctuation prediction, a complementary direction to our controlled comparison of single-task heads. More recently, \citet{huyen2026acoustic} integrate punctuation and capitalization directly into an end-to-end Vietnamese ASR output vocabulary for edge deployment, eliminating a separate restoration module but coupling formatting to acoustic-model training. In contrast, the large-scale ASR data pipeline of \citet{vu2026vietnamese} uses a separately fine-tuned text model to restore punctuation and capitalization, showing that standalone restoration remains relevant alongside end-to-end alternatives. Across these six studies, no work simultaneously isolates head choice from backbone, scale, and optimization budget, which is the gap this benchmark addresses.

All encoders build on the Transformer \citep{vaswani2017attention}. BERT provides the bidirectional fine-tuning baseline \citep{devlin2019bert}, XLM-R a strong multilingual encoder \citep{conneau2020xlmr}, ELECTRA a contrasting replaced-token-discrimination objective \citep{clark2020electra}, and PhoBERT a monolingual Vietnamese encoder plus the large PhoPunct backbone \citep{nguyen2020phobert}. Together they separate backbone effects from prediction-head effects.

At the prediction level, linear-chain CRFs explicitly score transitions between adjacent labels \citep{lafferty2001crf}, and Bi-LSTM--CRF models combine recurrent context with global decoding \citep{huang2015bilstmcrf}. These mechanisms are well motivated for outputs with strong local constraints. Punctuation, however, does not follow a rigid BIO-style grammar, and much of its context may already be encoded by self-attention. Moreover, transfer-learning research shows that newly initialized and pretrained layers can require different update scales \citep{howard2018ulmfit}. The present work therefore compares Plain, Bi-LSTM, CRF, and Bi-LSTM--CRF heads under one shared budget, making the distinction between architectural capacity and architecture-specific tuning explicit.

Table~\ref{tab:related-comparison} summarizes the methodological gap across the Vietnamese studies. Our benchmark alone combines a complete matched head grid, a uniform optimization budget, and 22.1M-token scale. Because the gap concerns architectural control and data scale jointly, Section~\ref{sec:dataset} establishes the corpus size and class distribution, while Section~\ref{sec:method} defines the controlled grid; together they support the fixed-budget comparison in Section~\ref{sec:results}.

\section{Dataset}
\label{sec:dataset}

\subsection{Data source and scale}

We use the News and Novels corpora originally released by \citet{pham2020vietnamese}; the files we train on are identical to the ones they publicly distributed. \citet{tran2021efficient} also builds on this corpus, although that paper reports substantially smaller sentence counts (Table~\ref{tab:related-comparison}) than the full files inherited here; we report the discrepancy without inferring its cause. News consists of Vietnamese journalistic text, whereas Novels contains literary prose. We retain the original core train, validation, and test partitions and exclude synthetic augmentation. The distributed data use a CoNLL-style format with one labeled token per line; our contribution is a controlled benchmark on these inherited splits rather than a newly annotated corpus. Table~\ref{tab:data-scale} reports the resulting token counts.

\begin{table}[t]
    \centering
    \small
    \begin{tabular}{llr}
        \toprule
        Domain & Split & Tokens \\
        \midrule
        News & Train & 11,939,762 \\
             & Validation & 4,003,912 \\
             & Test & 3,958,621 \\
             & Total & \textbf{19,902,295} \\
        \midrule
        Novels & Train & 1,350,500 \\
               & Validation & 358,791 \\
               & Test & 526,011 \\
               & Total & \textbf{2,235,302} \\
        \midrule
        Overall & & \textbf{22,137,597} \\
        \bottomrule
    \end{tabular}
    \caption{Dataset scale by domain and split.}
    \label{tab:data-scale}
\end{table}

\subsection{Domain characteristics}
\label{sec:domain-characteristics}

The two domains create different restoration conditions. News uses comparatively formal, standardized prose and supplies many repetitions of common punctuation contexts. Novels contain more dialogue, quotations, fragments, and variable sentence boundaries, increasing reliance on local discourse cues and shifting the label distribution. The domain split therefore tests whether conclusions about prediction heads remain stable across writing styles.

\subsection{Inherited preprocessing and task construction}

We use the processed corpus supplied by the prior pipeline rather than reconstructing raw documents. Each record pairs a lexical token with its following punctuation label; punctuation is withheld from the input and the original splits are preserved. During input construction, each word's label is attached to its first-subword representation (Section~\ref{sec:task-formulation}). We check UTF-8 readability, label validity, token--label alignment, empty examples, and split counts, but cannot audit semantic duplication without source-document identifiers.

\begin{figure}[t]
  \centering
  \makebox[\columnwidth][c]{%
    \hspace*{-0.1cm}%
    \includegraphics[width=\columnwidth]{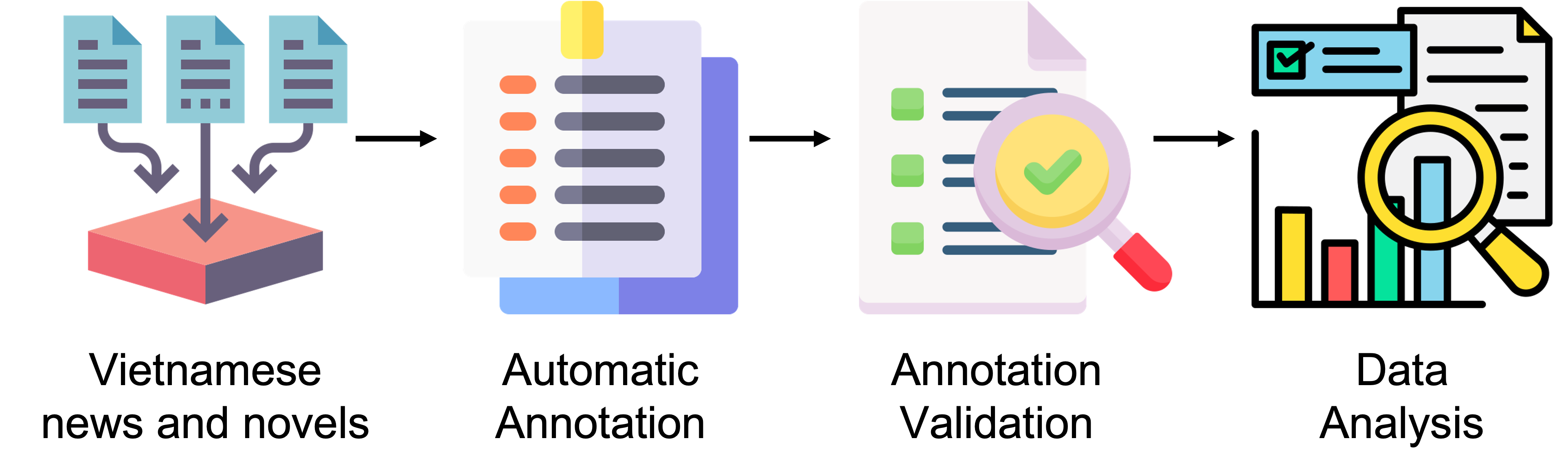}%
    \hspace*{0.1cm}%
  }
  \caption{Validation workflow for the inherited corpus. The study checks the distributed format and label alignment; it does not claim to recreate or re-annotate the original data.}
  \label{fig:validation-pipeline}
\end{figure}

Figure~\ref{fig:validation-pipeline} summarizes how the inherited data are prepared for the benchmark. The workflow deliberately begins with the distributed News and Novels corpus rather than raw-source collection, then applies format and alignment checks before class-distribution analysis.

\subsection{Class distribution}
\label{sec:label-imbalance}

\begin{figure}[t]
  \centering
  \includegraphics[width=\columnwidth]{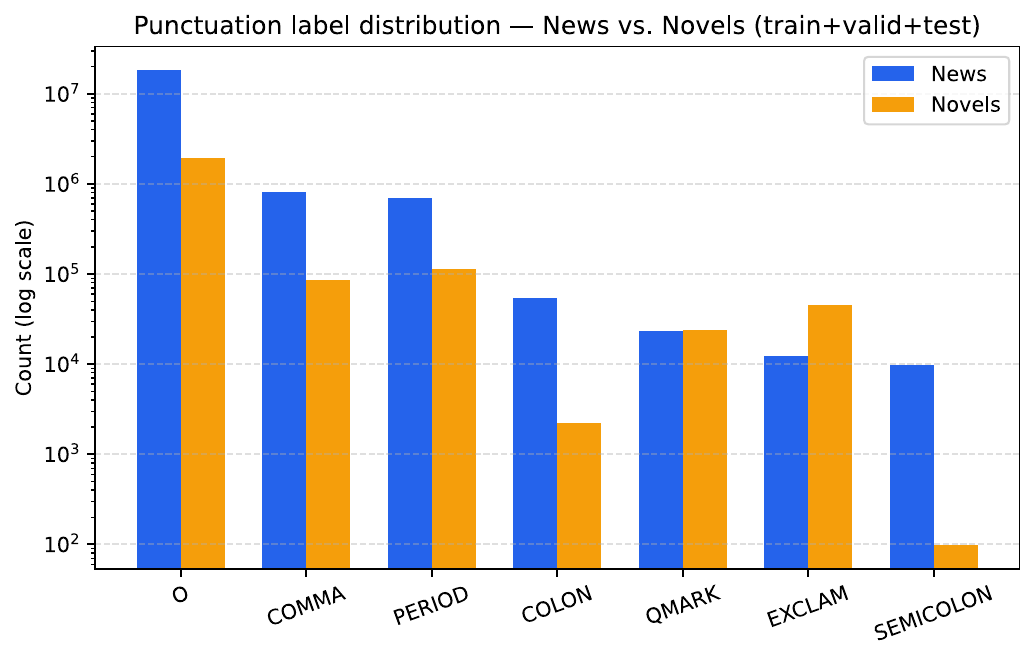}
  \caption{Punctuation-label counts by domain over the complete train, validation, and test splits. Counts are displayed on a logarithmic scale.}
  \label{fig:label-distribution}
\end{figure}

The label inventory consists of \textsc{O}, \textsc{Period}, \textsc{Comma}, \textsc{Colon}, \textsc{QMark}, \textsc{Exclam}, and \textsc{Semicolon}. Figure~\ref{fig:label-distribution} shows the aggregate distribution over train, validation, and test data. The log scale reveals several orders of magnitude between the no-punctuation class and the rarest marks. The imbalance is domain dependent: Novels contains substantially fewer colons and semicolons than common marks, despite having relatively more exclamation marks than News.

The Novels test report contains 320 colons but only 6 semicolons. Consequently, a few decisions can determine an entire per-class F1 score: both domain-leading systems later obtain 0.00 semicolon F1 (Section~\ref{sec:rare-results}). Macro F1 intentionally exposes this instability because every punctuation class contributes equally, unlike micro F1, which would be dominated by \textsc{O} and other frequent labels.

\section{Method}
\label{sec:method}

\subsection{Task formulation}
\label{sec:task-formulation}

Let $w_1,\ldots,w_n$ be an unpunctuated Vietnamese word sequence. The task is to predict one label $y_i\in\mathcal{Y}$ for every word $w_i$. The inventory $\mathcal{Y}$ contains \textsc{O}, \textsc{Period}, \textsc{Comma}, \textsc{Colon}, \textsc{Qmark}, \textsc{Exclam}, and \textsc{Semicolon}; \textsc{O} means that no punctuation follows the word, whereas each other label denotes the mark that immediately follows it.

The input is the lowercased, punctuation-stripped word sequence derived from a News or Novels text. The output is a sequence $y_1,\ldots,y_n$, with exactly one word-level label for each input word.

A pretrained tokenizer may divide a word into multiple subwords, but the label inventory remains defined at the word level. Every architecture must therefore produce exactly one prediction per original word, regardless of its subword count. We enforce this common alignment requirement by retaining the final-layer representation of each word's first subword; Section~\ref{sec:architectures} details this gather operation and the subsequent prediction heads.

\subsection{Architectures}
\label{sec:architectures}

\subsubsection{Backbones}

We evaluate four base-sized pretrained Transformer encoders selected to span distinct pretraining paradigms. This diversity tests whether conclusions about prediction heads persist across multilingual and Vietnamese-specific encoder families.

\nolinkurl{bert-base-multilingual-cased} (mBERT) uses masked-language-model pretraining across 104 languages, including Vietnamese, without a Vietnamese-specific vocabulary or corpus \citep{devlin2019bert}. We include it as a non-specialized multilingual reference against which the benefits of Vietnamese-focused or larger-scale multilingual pretraining can be assessed.

\nolinkurl{xlm-roberta-base} (XLM-R) applies RoBERTa-style masked-language modeling to 2.5TB of CommonCrawl text across 100 languages \citep{conneau2020xlmr}. It provides a stronger multilingual comparison for examining the effect of substantially greater pretraining scale and data coverage than mBERT.

vELECTRA (\nolinkurl{FPTAI/}\allowbreak\nolinkurl{velectra-base-discriminator-cased}) is a monolingual Vietnamese encoder based on ELECTRA's replaced-token-detection objective rather than masked-language modeling \citep{clark2020electra}. We include it to test whether discriminative pretraining transfers differently to punctuation restoration than the masked-language-modeling encoders.

\nolinkurl{NlpHUST/vibert4news-base-cased} (ViBERT) is a monolingual Vietnamese BERT pretrained on news-domain text. It serves as a domain-matched reference for the News split and tests whether specialized pretraining remains advantageous when the prediction head is held constant.

PhoBERT-large is a monolingual Vietnamese encoder pretrained on a 20GB corpus with the RoBERTa objective \citep{nguyen2020phobert}. It supplies the backbone only for PhoPunct rather than the matched grid because comparing a large model directly with the four base-sized encoders would confound backbone family with model size.

\subsubsection{Prediction heads}

Each base-sized backbone is paired with Plain, Bi-LSTM, CRF, and Bi-LSTM--CRF heads. Following the alignment requirement in Section~\ref{sec:task-formulation}, the backbone representation $h_i$ for word $w_i$ is gathered from the final layer at that word's first-subword position.

For a Plain model, a linear layer maps $h_i$ directly to label logits:
\begin{equation}
    z_i = W h_i + b.
\end{equation}
The Plain model minimizes token-level cross-entropy,
\begin{equation}
    \mathcal{L}_{\mathrm{CE}}
    = -\sum_{i=1}^{n}\log \operatorname{softmax}(z_i)_{y_i}.
\end{equation}
This configuration tests how much of the task is already solved by Transformer context without additional trainable sequence modeling.

For Bi-LSTM variants, the Transformer representations are transformed in both sequence directions:
\begin{align}
    \overrightarrow{s}_i
      &= \operatorname{LSTM}_{f}(h_i,\overrightarrow{s}_{i-1}), \\
    \overleftarrow{s}_i
      &= \operatorname{LSTM}_{b}(h_i,\overleftarrow{s}_{i+1}), \\
    g_i &= [\overrightarrow{s}_i;\overleftarrow{s}_i].
\end{align}
The recurrent head can reinforce local word order and dialogue-like patterns, although it may be redundant when self-attention already encodes the relevant context.

The emission vector is computed from $r_i=h_i$ for CRF-only models and from $r_i=g_i$ for Bi-LSTM models:
\begin{equation}
    e_i=W_r r_i+b_r.
\end{equation}
For CRF variants, transition scores $A$ define the sequence score and normalized probability
\begin{align}
 s(\mathbf{y},\mathbf{r})
   &= \sum_{i=1}^{n}\left[A_{y_{i-1},y_i}+e_{i,y_i}\right], \\
 P(\mathbf{y}\mid\mathbf{r})
   &= \frac{\exp s(\mathbf{y},\mathbf{r})}
   {\sum_{\mathbf{y}'\in\mathcal{Y}^{n}}
    \exp s(\mathbf{y}',\mathbf{r})}.
\end{align}
CRF variants minimize $\mathcal{L}_{\mathrm{CRF}}=-\log P(\mathbf{y}\mid\mathbf{r})$, and inference selects the highest-scoring complete label sequence. Globally normalized decoding allows adjacent labels to influence one another, although punctuation dependencies are less rigid than BIO constraints.

The Bi-LSTM--CRF configuration combines the recurrent contextualization in Equations~(3)--(5) with the structured decoding in Equations~(7)--(8). It is therefore the most elaborate prediction head in the matched grid.

\begin{figure}[t]
\centering
\begin{tikzpicture}[
    >=Latex,
    node distance=0.45cm,
    every text node part/.style={align=center},
    basebox/.style={
        rectangle,
        rounded corners=3.5pt,
        line width=0.8pt,
        inner xsep=5pt,
        inner ysep=4pt,
        text width=0.96\columnwidth,
        font=\small\linespread{0.95}\selectfont
    },
    inputbox/.style={basebox, fill=boxbluebg, draw=boxblueborder},
    prepbox/.style={basebox, fill=boxredbg, draw=boxredborder},
    backbonebox/.style={basebox, fill=boxyellowbg, draw=boxyellowborder},
    classifierbox/.style={basebox, fill=boxgreenbg, draw=boxgreenborder},
    optionalbox/.style={basebox, fill=boxpurplebg, draw=boxpurpleborder, dashed, line width=0.9pt},
    flowarrow/.style={->, line width=0.85pt, color=arrowgray}
]

    \node[inputbox] (input) {
        \textbf{Input Words} \\
        \footnotesize \textcolor{darkgray}{Raw Vietnamese tokens: $\mathbf{w} = (w_1, w_2, \dots, w_N)$}
    };

    \node[prepbox, below=of input] (tok) {
        \textbf{Tokenization + First-Subword Gather} \\
        \footnotesize \textcolor{darkgray}{Word-to-subword alignment \& representation pooling}
    };

    \node[backbonebox, below=of tok] (backbone) {
        \textbf{Transformer Backbone} $(d \in \{768, 1024\})$ \\
        \footnotesize \textcolor{darkgray}{PhoBERT-base / large, ViDeBERTa, XLM-RoBERTa, etc.}
    };

    \node[optionalbox, below=of backbone] (bilstm) {
        \textbf{Bi-LSTM Layer} $(128 \times 2)$ \\
        \footnotesize \textcolor{darkgray}{Sequential context aggregation}
    };

    \node[classifierbox, below=of bilstm] (linear) {
        \textbf{Linear Classifier} \\
        \footnotesize \textcolor{darkgray}{Dense projection to punctuation space ($K=7$ classes)}
    };

    \node[optionalbox, below=of linear] (crf) {
        \textbf{CRF Decoding} \\
        \footnotesize \textcolor{darkgray}{Viterbi sequence label transition modeling}
    };

    \node[inputbox, below=of crf] (output) {
        \textbf{Punctuation Sequence} \\
        \footnotesize \textcolor{darkgray}{Predicted target labels $\mathbf{y} \in \{\text{O, comma, period, \dots}\}^N$}
    };

    \draw[flowarrow] (input) -- (tok);
    \draw[flowarrow] (tok) -- (backbone);
    \draw[flowarrow] (backbone) -- (bilstm);
    \draw[flowarrow] (bilstm) -- (linear);
    \draw[flowarrow] (linear) -- (crf);
    \draw[flowarrow] (crf) -- (output);

\end{tikzpicture}
\caption{Matched prediction-head framework (dashed boxes denote optional components). Each backbone produces one contextual representation per word through first-subword gathering. The Plain, Bi-LSTM, CRF, and Bi-LSTM--CRF configurations use the same classifier-module parameterization across the four base-sized backbones; PhoBERT-large supplies the backbone for PhoPunct.}
\label{fig:architecture-framework}
\end{figure}
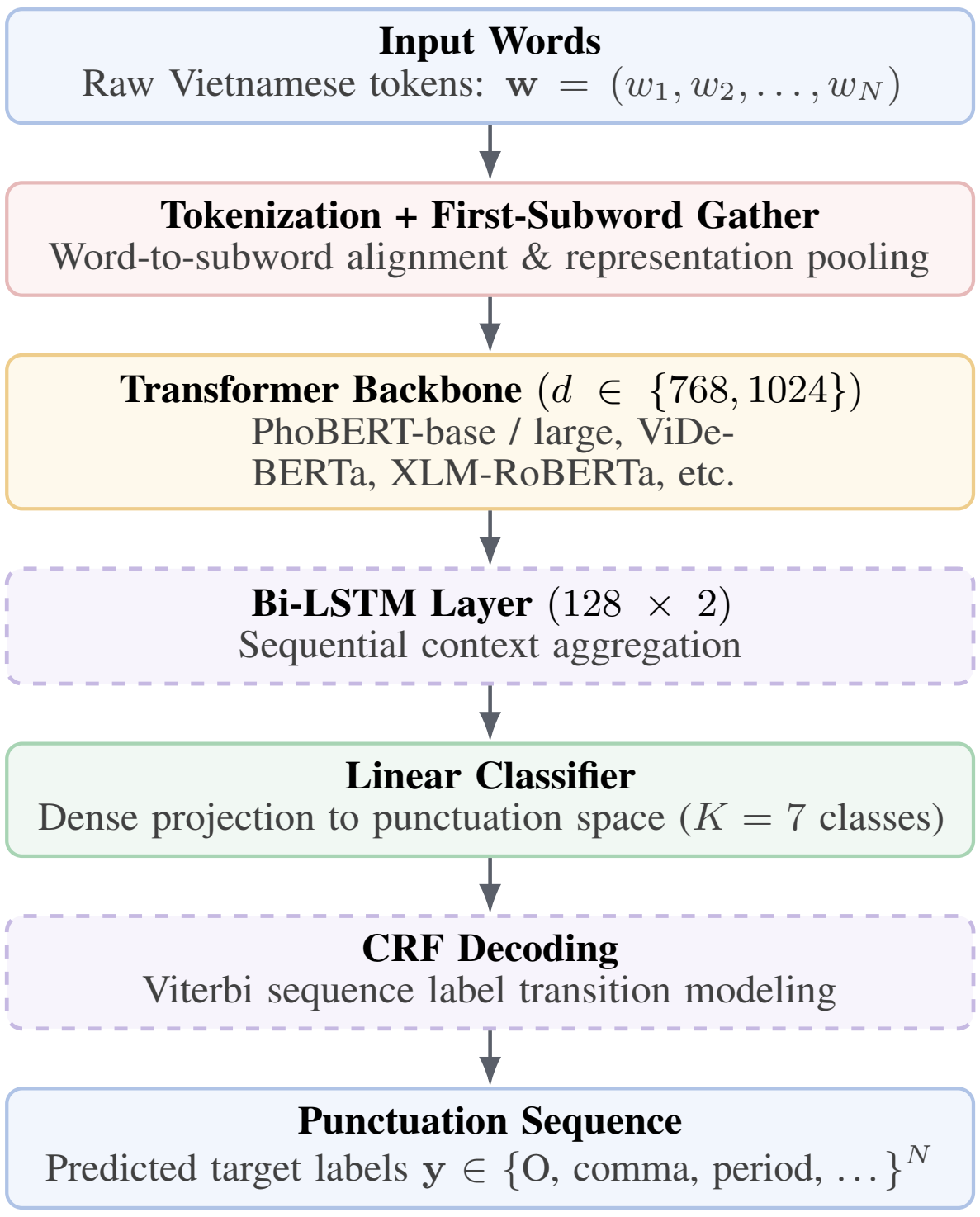

Figure~\ref{fig:architecture-framework} illustrates the shared word-aligned pipeline and the optional recurrent and structured-decoding components.

\subsubsection{Matched grid and PhoPunct}

The matched grid isolates prediction-head effects rather than searching separately for each backbone's best architecture. Within a backbone, the encoder, data pipeline, and training budget remain fixed, so macro-F1 changes can be attributed directly to the selected head. The classifier follows a common parameterization---dropout, an optional Bi-LSTM, dropout and nonlinear projection, and optional CRF decoding---instead of granting any backbone a specially tuned module. Crossing four heads with four base-sized backbones yields 16 controlled baselines.

PhoPunct is the seventeenth architecture and combines PhoBERT-large with the full Bi-LSTM--CRF head in Equations~(3)--(8). It is evaluated as a representative complete system rather than as another point in the matched base-model grid.

\subsection{Training setup}

All models are trained on the inherited training split, selected on validation macro F1, and evaluated once on the corresponding test split. We use AdamW \citep{loshchilov2019adamw}, mixed precision, and the shared configuration in Table~\ref{tab:hyperparameters}. Training ran on two NVIDIA RTX 3090-class GPUs; software versions and transient host identifiers do not vary across the compared heads.

\begin{table}[t]
    \centering
    \small
    \begin{tabular}{ll}
        \toprule
        Hyperparameter & Value \\
        \midrule
        Train/eval batch size & 16 / 32 \\
        Learning rate & $2\times10^{-5}$ \\
        Optimizer & AdamW \\
        Warmup ratio & 0.06 \\
        Epochs & 8 \\
        Maximum length & 256 subwords \\
        Precision & FP16 \\
        Gradient clipping & 1.0 \\
        Bi-LSTM hidden size & 128 per direction \\
        \bottomrule
    \end{tabular}
    \caption{Training hyperparameters shared by all architectures.}
    \label{tab:hyperparameters}
\end{table}

The single learning rate is applied to both pretrained and newly initialized parameters, with no architecture-specific search. This makes the experiment a comparison under equal optimization budgets rather than an estimate of each architecture's separately tuned optimum; Section~\ref{sec:discussion-heads} discusses the resulting tradeoff.

\subsection{Evaluation}

We report macro F1 over punctuation classes for both test domains. Equal class weighting exposes rare-class failures that micro F1 would conceal. Because preprocessing, splits, evaluation, and training schedules are fixed, within-backbone differences provide direct evidence about prediction-head behavior.

\section{Results}
\label{sec:results}

\subsection{Overall comparison}

\begin{figure*}[t]
  \centering
  \includegraphics[width=0.96\textwidth]{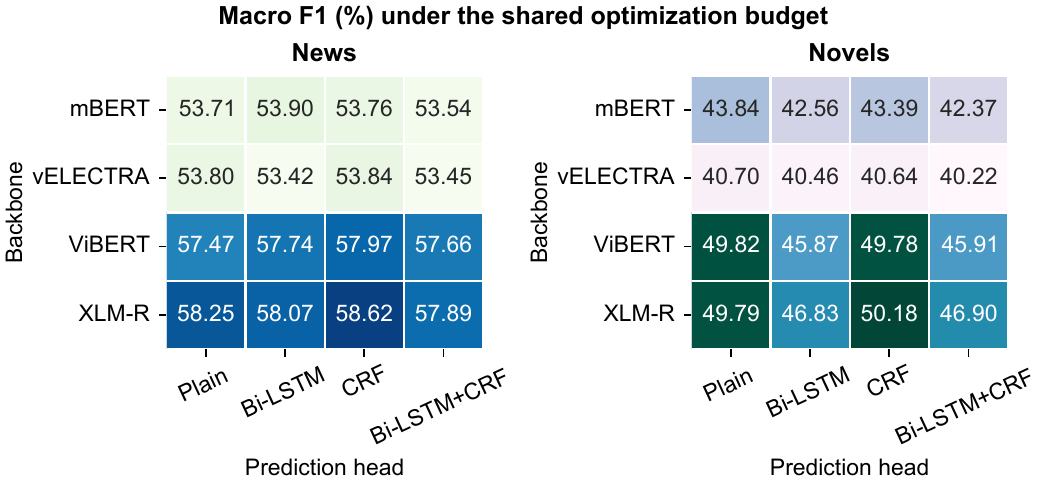}
  \caption{Macro F1 for the controlled backbone--head grid. The same four head configurations are compared within each base-sized backbone; PhoPunct is reported separately in Table~\ref{tab:full_results} because it uses PhoBERT-large.}
  \label{fig:overall-visuals}
\end{figure*}

Our first question is whether the 22.1M-token benchmark yields one stable architecture ranking across domains. Table~\ref{tab:full_results} shows that it does not: PhoPunct leads News at 61.35 macro F1, whereas XLM-R+CRF leads Novels at 50.18, ahead of PhoPunct's 48.97. The large Vietnamese encoder therefore provides a clear News advantage but not a domain-independent one.

\begin{table}[t]
    \centering
    \footnotesize
    \setlength{\tabcolsep}{3pt}
    \renewcommand{\arraystretch}{0.95}
    \begin{tabularx}{\columnwidth}{@{}r>{\raggedright\arraybackslash}Xrr@{}}
        \toprule
        ID & Model & News & Novels \\
        \midrule
        1 & mBERT & 53.71 & 43.84 \\
        2 & vELECTRA & 53.80 & 40.70 \\
        3 & ViBERT & 57.47 & 49.82 \\
        4 & XLM-R & 58.25 & 49.79 \\
        5 & mBERT + Bi-LSTM & 53.90 & 42.56 \\
        6 & vELECTRA + Bi-LSTM & 53.42 & 40.46 \\
        7 & ViBERT + Bi-LSTM & 57.74 & 45.87 \\
        8 & XLM-R + Bi-LSTM & 58.07 & 46.83 \\
        9 & mBERT + CRF & 53.76 & 43.39 \\
        10 & vELECTRA + CRF & 53.84 & 40.64 \\
        11 & ViBERT + CRF & 57.97 & 49.78 \\
        12 & XLM-R + CRF & 58.62 & \textbf{50.18} \\
        13 & mBERT + Bi-LSTM + CRF & 53.54 & 42.37 \\
        14 & vELECTRA + Bi-LSTM + CRF & 53.45 & 40.22 \\
        15 & ViBERT + Bi-LSTM + CRF & 57.66 & 45.91 \\
        16 & XLM-R + Bi-LSTM + CRF & 57.89 & 46.90 \\
        17 & PhoPunct (proposed) & \textbf{61.35} & 48.97 \\
        \bottomrule
    \end{tabularx}
    \caption{Macro F1 (\%) across all 17 architectures. Bold indicates the best result per domain.}
    \label{tab:full_results}
\end{table}

Figure~\ref{fig:overall-visuals} reorganizes the same evidence as a matched $4\times4$ grid. XLM-R is the strongest base-sized backbone and CRF is its best head in both domains, but the gains over Plain are only 0.37 points on News and 0.39 on Novels. The central result is consequently comparative rather than leaderboard-oriented: scale makes domain effects visible, while the grid identifies whether they arise from the encoder or the head.

\subsection{Plain versus Bi-LSTM--CRF models}

Our second question is whether the most elaborate head consistently improves a fixed encoder. Table~\ref{tab:plain_vs_crf} answers negatively: Plain matches or exceeds Bi-LSTM--CRF in seven of eight comparisons. The exception is ViBERT on News ($-0.19$), while the largest Plain advantages occur on Novels for ViBERT ($+3.91$) and XLM-R ($+2.89$). Under the shared budget, additional sequence modeling is therefore not a reliable substitute for a strong contextual representation. Figure~\ref{fig:phopunct-architecture} summarizes the full PhoPunct path.

\begin{table}[t]
    \centering
    \small
    \setlength{\tabcolsep}{3pt}
    \begin{tabular*}{\columnwidth}{@{\extracolsep{\fill}}llrrr@{}}
        \toprule
        Backbone & Domain & Plain & B+CRF & $\Delta$ \\
        \midrule
        mBERT & News & 53.71 & 53.54 & $+0.17$ \\
        mBERT & Novels & 43.84 & 42.37 & $+1.47$ \\
        vELECTRA & News & 53.80 & 53.45 & $+0.35$ \\
        vELECTRA & Novels & 40.70 & 40.22 & $+0.48$ \\
        ViBERT & News & 57.47 & 57.66 & $-0.19$ \\
        ViBERT & Novels & 49.82 & 45.91 & $+3.91$ \\
        XLM-R & News & 58.25 & 57.89 & $+0.36$ \\
        XLM-R & Novels & 49.79 & 46.90 & $+2.89$ \\
        \bottomrule
    \end{tabular*}
    \caption{Matched Plain and Bi-LSTM--CRF (B+CRF) results. $\Delta$ is Plain minus B+CRF macro F1.}
    \label{tab:plain_vs_crf}
\end{table}

{\setlength{\intextsep}{4pt}
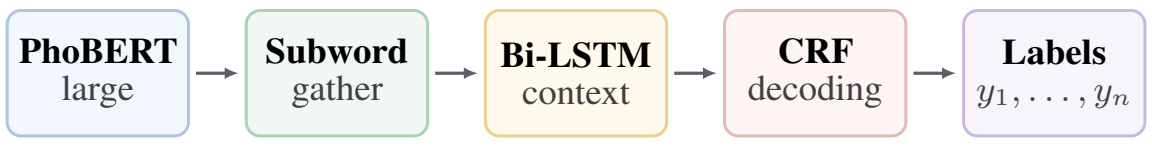
\begin{figure}[t]
\centering
\resizebox{0.98\columnwidth}{!}{%
\begin{tikzpicture}[
    >=Latex,
    node distance=0.48cm,
    every text node part/.style={align=center},
    flowbox/.style={
        rectangle,
        rounded corners=3.5pt,
        line width=0.75pt,
        inner xsep=3pt,
        inner ysep=4pt,
        minimum height=1.1cm,
        text width=1.38cm,
        align=center,
        font=\small\linespread{0.92}\selectfont
    },
    bbstyle/.style={flowbox, fill=boxbluebg,   draw=boxblueborder},
    sgstyle/.style={flowbox, fill=boxgreenbg,  draw=boxgreenborder},
    lstmstyle/.style={flowbox, fill=boxyellowbg, draw=boxyellowborder},
    crfstyle/.style={flowbox, fill=boxredbg,    draw=boxredborder},
    lblstyle/.style={flowbox, fill=boxpurplebg, draw=boxpurpleborder},
    flowarrow/.style={
        -{Latex[length=3.5pt, width=3pt]},
        line width=0.8pt,
        color=arrowgray,
        shorten <= 1.5pt,
        shorten >= 1.5pt
    }
]

    \node[bbstyle] (phobert) {
        \textbf{PhoBERT} \\
        \footnotesize \textcolor{darkgray}{large}
    };

    \node[sgstyle, right=of phobert] (gather) {
        \textbf{Subword} \\
        \footnotesize \textcolor{darkgray}{gather}
    };

    \node[lstmstyle, right=of gather] (bilstm) {
        \textbf{Bi-LSTM} \\
        \footnotesize \textcolor{darkgray}{context}
    };

    \node[crfstyle, right=of bilstm] (crf) {
        \textbf{CRF} \\
        \footnotesize \textcolor{darkgray}{decoding}
    };

    \node[lblstyle, right=of crf] (labels) {
        \textbf{Labels} \\
        \footnotesize \textcolor{darkgray}{$y_1, \dots, y_n$}
    };

    \draw[flowarrow] (phobert) -- (gather);
    \draw[flowarrow] (gather)  -- (bilstm);
    \draw[flowarrow] (bilstm)  -- (crf);
    \draw[flowarrow] (crf)     -- (labels);

\end{tikzpicture}%
}
\caption{PhoPunct architecture and decoding flow.}
\label{fig:phopunct-architecture}
\end{figure}}

\subsection{Rare punctuation classes}
\label{sec:rare-results}

\begin{figure*}[t]
    \centering
    \includegraphics[width=0.94\textwidth]{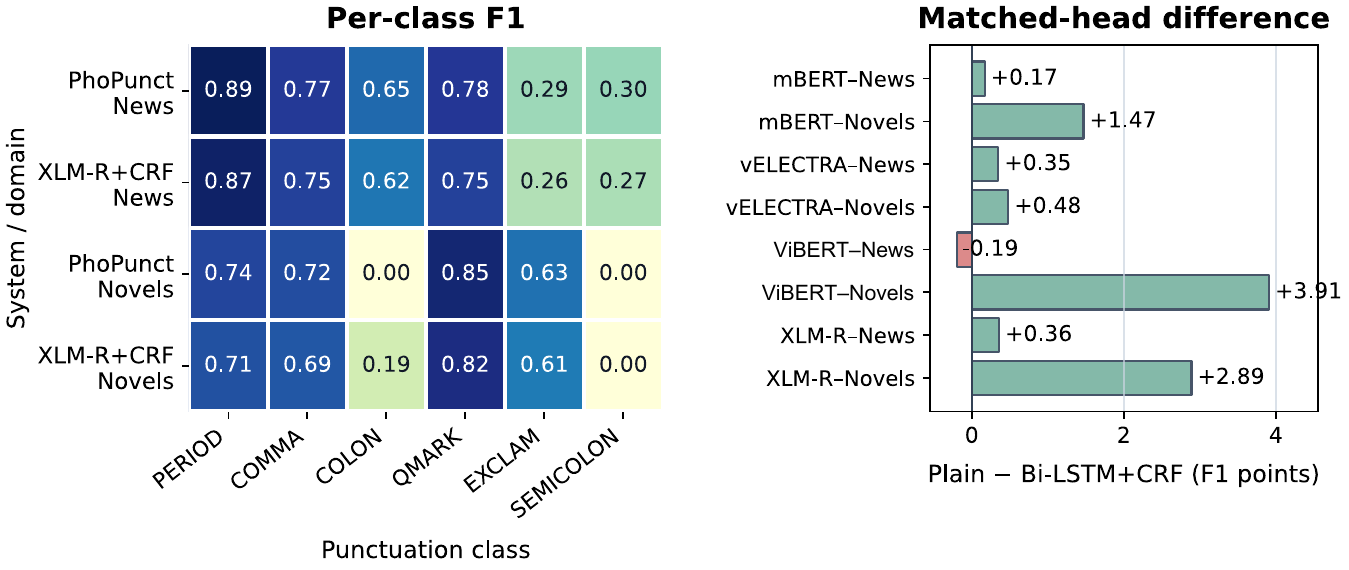}
    \caption{Diagnostic evidence. Left: per-class F1 for the two principal systems by domain, excluding \textsc{O}. Right: Plain minus Bi-LSTM--CRF macro F1; positive values favor Plain.}
    \label{fig:diagnostic-comparison}
\end{figure*}

Figure~\ref{fig:diagnostic-comparison} connects the aggregate findings to individual classes. PhoPunct is stronger across displayed News classes, but on Novels XLM-R+CRF recovers colon signal (0.19 versus 0.00). Neither system predicts any of the six semicolon cases correctly. Thus, the large corpus supports broad architectural comparison, yet extreme tail support still limits what aggregate macro F1 can establish about the rarest marks.

\section{Discussion}

\subsection{Why do additional sequence layers fail to help?}
\label{sec:discussion-heads}

The matched grid turns a broad ranking into a controlled diagnosis. A single $2\times10^{-5}$ rate suits gentle Transformer fine-tuning but may under-train randomly initialized Bi-LSTM and CRF parameters in eight epochs. Moreover, punctuation has weaker local transition constraints than BIO tagging, while the Transformer already supplies bidirectional context. The added head can therefore increase optimization burden without adding commensurate information, especially on Novels. This interpretation concerns the shared budget; it does not claim that separately tuned recurrent or structured heads are ineffective.

\subsection{Architecture versus tuning}
\label{sec:architecture-tuning}

The benchmark deliberately measures equal-budget behavior, not each model's tuned optimum. This distinction is scientifically useful because every within-backbone difference is attributable to the head under a reproducible protocol. A follow-up with separate encoder/head rates, longer schedules, validation-based stopping, and multiple seeds can test whether the observed gaps reflect optimization limits or redundant capacity.

\subsection{Domain effects}

PhoPunct's News lead and XLM-R+CRF's Novels lead show why both scale and domain coverage matter. Across all architectures, the lower performance on Novels reflects the domain's colloquial dialogues, quotations, and sentence fragments, which contrast with the standardized syntax of News and complicate boundary prediction. Formal News may align better with a large Vietnamese encoder, whereas dialogue-heavy Novels changes both sequence patterns and label frequencies. The colon contrast in Figure~\ref{fig:diagnostic-comparison} demonstrates a configuration-specific structured-decoding benefit, but the semicolon failure prevents a universal claim.

A closer comparison refines this interpretation. For XLM-R, CRF-only improves over Plain by almost the same amount on News ($+0.37$) and Novels ($+0.39$; Table~\ref{tab:full_results}), so CRF itself does not appear specifically advantaged by Novels. The domain-specific difference instead emerges with the full Bi-LSTM--CRF head: relative to Plain, it loses 3.91 points for ViBERT and 2.89 for XLM-R on Novels, whereas on News ViBERT gains 0.19 and XLM-R loses only 0.36 (Table~\ref{tab:plain_vs_crf}). The narrower question is therefore why inserting a recurrent layer before CRF is more costly on Novels.

One testable explanation extends the optimization argument in Section~\ref{sec:discussion-heads}. Novels contains 2.2M tokens compared with 19.9M in News (Table~\ref{tab:data-scale}), giving newly initialized Bi-LSTM parameters substantially less domain-specific training signal under the same eight-epoch, single-rate schedule. A CRF adds a small label-transition matrix above pretrained emissions, whereas a bidirectional LSTM introduces many more randomly initialized parameters; these parameters may remain undertrained on the smaller corpus. A second, non-exclusive hypothesis concerns structure: dialogue, quotations, and fragments are more frequent in Novels (Section~\ref{sec:domain-characteristics}), potentially making direct local label transitions more useful than recurrent context that requires more updates to learn. The present single-run protocol cannot distinguish data-volume from sequence-structure effects. Separate head learning rates, longer Novels training, and multiple seeds, as proposed in Section~\ref{sec:architecture-tuning}, would test these alternatives.

\section{Limitations}

The two inherited text domains do not represent conversational ASR, social media, regional varieties, or specialized language. Raw-document identifiers remain unavailable, preventing an independent audit of annotation decisions and semantic duplication. The equal-budget protocol uses one run per configuration and may disadvantage newly initialized heads, so small gaps require multi-seed confirmation. Finally, the six Novels semicolons are insufficient for stable class-level conclusions; targeted collection or class-sensitive training would address a different experimental question.

\section{Conclusion}

We presented a controlled evaluation of 17 Vietnamese punctuation-restoration architectures at the scale reported in Section~\ref{sec:dataset}. PhoPunct performs best on News, whereas XLM-R+CRF leads on Novels; in seven of eight matched comparisons, Plain matches or exceeds Bi-LSTM--CRF under the shared eight-epoch, single-rate budget. Thus, recurrent and structured heads do not automatically improve contextual encoders. Future work should separate encoder/head rates, test longer schedules and multiple seeds, and extend evaluation to conversational ASR and other Vietnamese domains.

\section*{Ethical Considerations}

The inherited News and Novels text may contain copyrighted material, social stereotypes, and uneven coverage of Vietnamese varieties; trained models may reproduce these patterns. We collect no sensitive annotations or demographic attributes, and any artifact release must respect the source licenses while documenting tokenization, filtering, and split handling. The intended use is academic text restoration and ASR post-processing, not surveillance, authorship profiling, or automated censorship; deployment requires domain-specific performance and legal review.

\clearpage

\bibliography{custom}

\end{document}